\documentclass[final,5p,times]{elsarticle}
\usepackage{graphicx}
\usepackage{subcaption}
\usepackage{epsfig}
\usepackage{siunitx}
\usepackage{booktabs}
\usepackage[T1]{fontenc}
\usepackage{multirow} 
\usepackage{tabularx}
\usepackage{array}
\usepackage[namelimits]{amsmath} 
\usepackage{amssymb}             
\usepackage{amsfonts}            
\usepackage{mathrsfs} 
\usepackage{float}
\usepackage{cases}
\usepackage{wrapfig}
\usepackage{hyphenat}
\usepackage{multicol}
\usepackage{caption}
\usepackage[labelfont=bf, textfont=normalfont, justification=raggedright, singlelinecheck=false]{caption}
\usepackage{graphicx}
\usepackage{subcaption}
\usepackage{amssymb}

\journal{arxiv}

\begin{document}
\begin{frontmatter}



\title{Capsule Endoscopy Image Enhancement for Small Intestinal Villi Clarity\tnoteref{13}}


\tnotetext[13]{This work was supported in part by the National Natural Science Foundation of China (No. 62172190), the "Double Creation" Plan of Jiangsu Province (Certificate: JSSCRC2021532) and the "Taihu Talent-Innovative Leading Talent" Plan of Wuxi City(Certificate Date: 202110). }
\author[label1]{Shaojie Zhang}
\ead{7213107006@stu.jiangnan.edu.cn}
\author[label1,label2]{Yinghui Wang\corref{cor1}}
\ead{wangyh@jiangnan.edu.cn}
\author[label1]{Peixuan Liu\corref{cor1}}
\ead{362342276@qq.com}
\author[label1]{Yukai Wang}
\ead{ericwangyk22@163.com}
\author[label3]{Liangyi Huang}
\ead{lhuan139@asu.edu}
\author[label4]{Mingfeng Wang}
\ead{mingfeng.wang@brunel.ac.uk}
\author[label5]{Ibragim. Atadjanov}
\ead{ibragim.atadjanov@gmail.com}
\cortext[cor1]{Corresponding author}
\affiliation[label1]{organization={ School of Artificial Intelligence and Computer Science, Jiangnan University},
            addressline={1800 Li Lake Avenue},
            city={wuxi},
            postcode={214122},
            state={Jiangsu},
            country={PR China}}
\affiliation[label2]{organization={ Engineering Research Center of Intelligent Technology for Healthcare, Ministry of Education},
            addressline={1800 Li Lake Avenue},
            city={wuxi},
            postcode={214122},
            state={Jiangsu},
            country={PR China}}
 \affiliation[label3]{organization={School of Computing and Augmented Intelligence, Arizona State University},
            addressline={1151 S Forest Ave},
            city={Tempe},
            postcode={8528},
            state={AZ},
            country={US}}
\affiliation[label4]{organization={Department of Mechanical and Aerospace Engineering, Brunel University},
            addressline={Kingston Lane},
            city={London},
            postcode={UB8 3PH},
            state={Middlesex},
            country={U.K}}    
\affiliation[label5]{organization={Tashkent University of Information Technologies named after al-Khwarizmi},
            addressline={ 108 Amir Temur Avenue},
            city={Tashkent},
            postcode={100084},
            country={Uzbekistan}}

\begin{abstract}
This paper presents, for the first time, an image enhancement method aimed at improving the clarity of small intestinal villi in Wireless Capsule Endoscopy (WCE) images. This method first separates the low-frequency and high-frequency components of small intestinal villi images using guided filtering. Subsequently, an adaptive light gain factor is generated based on the low-frequency component, and an adaptive gradient gain factor is derived from the convolution results of the Laplacian operator in different regions of small intestinal villi images. The obtained light gain factor and gradient gain factor are then combined to enhance the high-frequency components. Finally, the enhanced high-frequency component is fused with the original image to achieve adaptive sharpening of the edges of WCE small intestinal villi images. 
The experiments affirm that, compared to established WCE image enhancement methods, our approach not only accentuates the edge details of WCE small intestine villi images but also skillfully suppresses noise amplification, thereby preventing the occurrence of edge overshooting.
\end{abstract}



 \begin{keyword}


WCE images \sep Small intestinal villi images \sep Unsharp mask \sep Gain factor
\end{keyword}

\end{frontmatter}


\section{Introduction}
Wireless capsule endoscopy (WCE) allows for effective observation of the internal structures of the gastrointestinal tract while minimizing patient discomfort \cite{ref1}.
However, due to the limited illumination power of WCE and the complexity of the gastrointestinal environment, WCE images are often unclear, especially in the small intestine covered with tiny villi. 
An enlarged section of the real structure of the human small intestine is shown in Fig. 1 , revealing the surface of the small intestine covered with tiny villi responsible for nutrient absorption. 
Fig. 2 displays an actual image captured by WCE. A comparison shows that WCE images cannot effectively display the small intestinal villi structure and may even make it disappear, which is a common issue with WCE. 
The morphology of small intestinal villi serves as a crucial basis for diagnosing diseases such as indigestion and disorders in nutrient absorption. 
Enhancing techniques to highlight the microscopic villi structure and its edge details can assist doctors in better relying on villi morphology to observe small intestinal lesions. 
This enables a more precise identification of the location and severity of lesions, providing a more personalized treatment plan for patients \cite{ref2}. 
To the best of our knowledge, there are no articles in the literature reporting methods for enhancing WCE images to make villi more clearly visible.

\begin{figure}[!hptb]
	\setlength{\abovecaptionskip}{0.1cm}
    \begin{center}
    \includegraphics[width=3.5in]{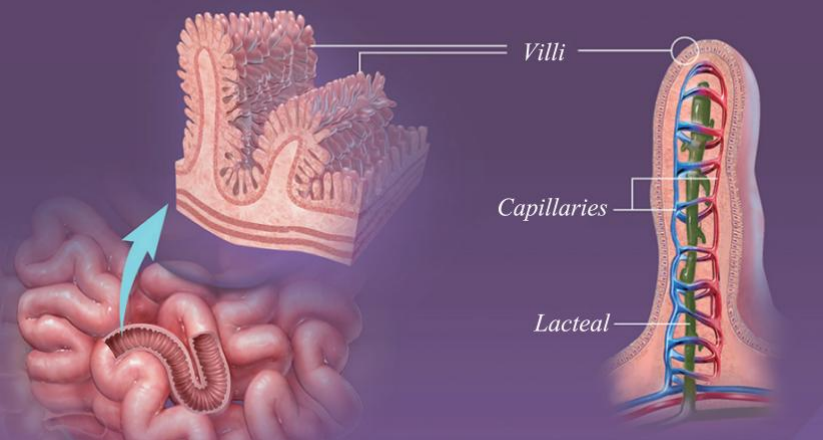}
	\setlength{\belowcaptionskip}{-0.5cm}
    \end{center}
\caption{Diagram of human small intestine structure.} 
\label{Fig1}
\end{figure}

\begin{figure}[!htbp]
	\centering
	\begin{subfigure}{0.24\linewidth}
		\centering
		\includegraphics[width=0.95\linewidth]{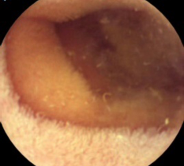}
		\label{21}%
	\end{subfigure}
	\centering
	\begin{subfigure}{0.24\linewidth}
		\centering
		\includegraphics[width=0.95\linewidth]{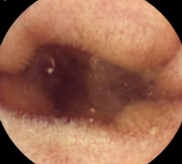}
		\label{22}%
	\end{subfigure}
	\centering
	\begin{subfigure}{0.24\linewidth}
		\centering
		\includegraphics[width=0.95\linewidth]{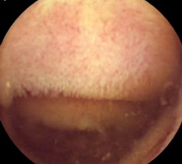}
		\label{23}%
	\end{subfigure}
	\centering
    \begin{subfigure}{0.24\linewidth}
		\centering
		\includegraphics[width=0.95\linewidth]{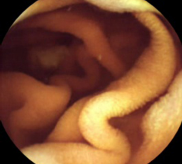}
		\label{24}%
	\end{subfigure}
	\caption{ WCE images of the small intestine with villin}
	\label{Fig2}
\end{figure}

So far, there have been numerous related methods for enhancing small intestine images. The commonly used classic 
methods can be classified into three categories: Histogram Equalization (HE) methods \cite{ref4,ref5,ref6,ref16}, 
Retine-based methods \cite{ref7,ref8,ref9,ref17,ref18} and Unsharping
mask (USM) methods\cite{ref10,ref11,ref12,ref13,ref14,ref19,ref20}, etc. 
Among them, the USM methods are the most advantageous approach in emphasizing image detail information. 
Methods based on HE enhance images by remapping the input image's gray levels using the probability distribution of grayscale levels. Nonetheless, these methods encounter problems like inadequate enhancement, excessive enhancement, and notable noise amplification in the resulting enhancements. 
The Retinex-based methods view an image as a combination of light and reflection components, enhancing the image by adjusting the corresponding components. However, the effect of highlighting image detail information using the Retinex-based methods is not very ideal.
Specifically, traditional USM methods use a fixed gain factor to enhance high-frequency components. Subsequently, some scholars improved the USM methods by proposing adaptive gain factors based on local image information \cite{ref21}.
Clearly, the aforementioned methods were not specifically designed for enhancing small intestinal villi. Moreover, our experiments have revealed \cite{ref22}  that they also do not achieve the desired effect of highlighting small intestinal villi.
In addition, in the past decade, many scholars have proposed methods for endoscope image enhancement based on USM \cite{ref11,ref14,ref15}. 
While USM methods have the potential to emphasize image details, these methods have not adequately considered the balance between detail enhancement and noise suppression. 
This imbalance results in issues such as edge overshooting and amplified noise in the enhanced results. Therefore, they are not directly suitable for the purpose of enhancing small intestine villi in endoscopic images.

\begin{figure*}[t]
    \centering
    \includegraphics[height=3in]{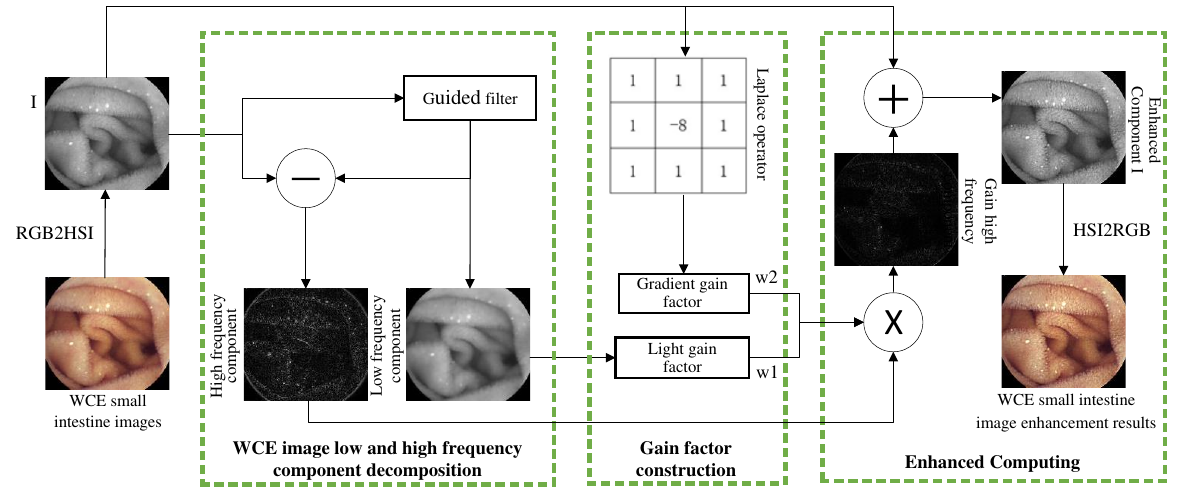}
\caption{Methodological Framework.}
    \label{Fig3}
\end{figure*}

To address this, this paper proposes an image enhancement method aimed at enhancing the clarity of small intestinal villi. 
The method aims to highlight the edge details of small intestinal villi in WCE images, achieving enhancement while suppressing excessive noise to prevent edge overshooting. 
The main innovations in this paper can be summarized as follows:

(1) We have proposed a gain factor construction method,
which relies on the light gain function and gradient gain 
function we construct. The light gain function combines light 
information to enhance high-frequency components while 
suppressing noise in darker areas. The gradient gain function 
enhances the high-frequency components of capsule endoscopy 
small intestine villi images based on gradient information. This 
enhances the fine details of small intestine villi while 
concurrently mitigating the occurrence of edge overshoot 
phenomena. 

(2) We have achieved an image-enhancement-based method 
for reproducing the small intestinal surface's tiny villi. This 
method provides medical professionals or in-body robotic 
systems with more opportunities for precise localization of 
lesions.

\section{RELATED WORK}
Three prevalent types of image enhancement methods include HE methods \cite{ref4,ref5,ref6,ref16}, Retine-based methods\cite{ref7,ref8,ref9,ref17,ref18}, and USM methods\cite{ref10,ref11,ref12,ref13,ref14,ref19,ref20}.

The HE methods\cite{ref4} partition the image histogram by identifying local minima and assigning specific grayscale ranges to each partition. Subsequently, these partitions are equalized separately to enhance the image. Reference \cite{ref17}, 
the HE methods are improved by introducing brightness constraints to achieve image detail enhancement while preserving brightness. 
However, methods based on HE typically lead to insufficient or excessive enhancement of fine details, making them unsuitable for highlighting the villi in small intestine images. 

The Retinex-based methods view a scene image as a product of reflection and illumination components in the human eye \cite{ref18}, 
enhancing the scene image by adjusting these components accordingly\cite{ref3}. Reference \cite{ref19}, the decomposed reflection component is directly regarded as the enhancement result, 
but it can lead to the occurrence of halos and a loss of naturalness. Afterwards, scholars proposed a variational model\cite{ref10} for estimating reflection and illumination components, 
adjusting and combining these components to reconstruct the enhanced effect. While the Retinex-based methods' effectiveness on fine details of microscopic villi structures is limited.

The USM methods primarily enhance image details and edge information by emphasizing the image's high-frequency content. 
Reference  \cite{ref20}, the Laplacian operator is utilized for image filtering to enhance the image, but this method has many parameters and involves heavy computation. Reference \cite{ref21}, 
the difference between the iterative median filtering and the original image is used to obtain details of the image. However, this approach results in halo artifacts along the edges, with limited enhancement in smoothly lit areas. 
Reference \cite{ref11}, a Gaussian filter is applied to WCE images to accentuate image details, but this process can lead to issues like overshooting and excessive noise amplification. 
Reference\cite{ref12}, a hybrid median filter\cite{ref23} calculates the median values for square, cross-shaped, and diagonal windows to replace the central pixel of the filtering window. 
However, this literature does not take into account the impact of noise during image enhancement.
Reference \cite{ref13}, the high-frequency components of a guided image are added to the low-frequency components of the original image under the control of a gain factor. 
This factor is obtained using an optimization function based on guided filtering\cite{ref24} and weighted guided filtering\cite{ref25}. Additionally, in \cite{ref13}introduces a convolutional neural network to derive the gain factor. 
However, due to the considerable difficulty in obtaining in vivo data, this method is not applicable to images in an in-body environment. Although these methods improve upon USM techniques by enhancing filters and gain factors to boost detail enhancement, they still suffer from issues like edge overshooting and noise amplification when applied to image enhancement. The main problem lies in how gain factors are configured.

To address this, this paper employs guided filtering\cite{ref24} to filter WCE small intestine villi images, obtaining low-frequency components with well-preserved edges and gradients. 
Constructing light gain functions based on the low-frequency components of different regions in WCE small intestine villi images to adaptively generate light gain factors. Furthermore, 
we create adaptive gradient gain factors by building gradient gain functions using the Laplacian operator's convolution results in various areas of WCE small intestine villi images. Ultimately, 
we combine the previously mentioned illumination gain factors and the gradient gain factors to create the adaptive gain coefficients needed for enhancing the Unsharp Masking (USM) method, thereby improving the USM approach. Our method effectively enhances the clarity of small intestine villi details while suppressing noise in darker regions and preventing edge overshooting.

\section{METHODOLOGY}
The image enhancement method framework for small intestine villi clarity in WCE is illustrated in Fig. 3.

\subsection{Low-High Frequency Component Decomposition}
The USM is a sharpening enhancement technology, and the calculation formula as follows.

\begin{equation}
	\label{eq1}
	ZI(x,y)= I(x,y)+k(I(x,y)-\overline{I}(x,y))
\end{equation}
 Where $ ZI(x,y) $ is the enhanced image, $I $ denotes the original image, $k$ stands for the gain coefficient, $\overline{I}(x,y) $ represents the image after low-pass filtering, which is the low-frequency component.

As can be observed, the essence of the USM technique involves subtracting the low-frequency component from the original image to obtain the high-frequency component \cite{ref16}. Then, the high-frequency component is amplified by multiplying it with a gain factor. Finally, the amplified high-frequency component is superimposed on the original image to enhance the image's edges. To achieve a more accurate high-frequency component, it is crucial to select a suitable low-pass filter that effectively preserves edges while capturing low-frequency components.

The guided filtering\cite{ref24} exhibits a linear relationship between the guided image and the output image as follows.
\begin{equation}
	\label{eq2}
	q_{i}=a _{m}G_{i}+b_{m},\forall i\in \omega _{m}
\end{equation}
Where $G $ represents the guided image, $ q $ denotes the output image, $i $ and $ m$ are pixel indices, $\omega _{m}$ stands for the square window located at position $m$, and $a _{m} $ and $b _{m} $ are the coefficients of the linear function at position $m$.

The coefficients $a _{m} $ and $b _{m} $ in refer to (2) are obtained by minimizing the linear cost function, as follows.

\begin{equation}
	\label{eq3}
	E(a_{m},b_{m})= \sum_{i\in \omega _{m}}^{}\left [ (a_{m}G_{i}+b_{m}-p_{i})^{2} +\varepsilon a_{m}^{2}\right ]
\end{equation}
Where $p $ represents the input image, and $\varepsilon $ is a regularization parameter used to prevent $a_{m}$ from becoming too large.

Taking the gradient on both sides of refer to (2), simultaneously reveals that the guided image has a similar gradient to the output image. 
This indicates that the guided filter exhibits good edge-preserving characteristics. 
The guided filter achieves excellent edge preservation by leveraging the linear relationship between the guidance image and the output image. 
Therefore, this paper initially converts the WCE small intestine image from the RGB color space to the HSI color space. 
Subsequently, a guided filter is employed to smooth and filter the I component of the WCE small intestine image in the HSI color space, thereby obtaining its low-frequency component.

\subsection{Gain Factor Construction}
This section encompasses two aspects: the construction of light gain factor and the construction of gradient gain factor.
\subsubsection{Light Gain Factor Construction}
Due to the complex in-body environment and limited illumination of the WCE, images exhibit areas that are either too dark or too bright, with dark regions being particularly noisy, as shown in Fig.2. 
Traditional USM methods do not account for the impact of light on WCE images, resulting in post-enhancement WCE images with noisy dark regions and insufficient detail enhancement in other areas.
 To address this, this paper incorporates light gain factors as part of the enhancement coefficient k in refer to (1). The light gain factors in different regions of WCE images obtained through a light gain function, as follows.

\begin{equation}
	\label{eq4}
	w1(x,y)=\begin{Bmatrix} 0.5sin(\bar{I}(x,y)\cdot \pi ,\bar{I}(x,y)< Mean)\\ sin(\bar{I}(x,y)\cdot \pi ),otherwise \end{Bmatrix}
\end{equation}
Where $(x,y)$ is the pixel coordinate of the image and $\bar{I}(x,y)$ is the Mean.

Since the light changes slowly and smoothly belongs to low-frequency information, the low-frequency information $\bar{I}(x,y)$ obtained by the guided filter is used as the light information of the WCE image in this paper. The light gain function is shown in Fig. 4. When the brightness of WCE image is less than the Mean, it indicates the dark areas of the WCE image. The method described in this paper acquires the illumination gain factor using a compressed sine function. Subsequently, this illumination gain factor is incorporated into the gain coefficient of refer to (1). This process enhances high-frequency information in dark areas while simultaneously suppressing noise in those regions. When the brightness of the WCE image surpasses the mean, a sine function is utilized to derive the light gain factor. From Fig. 4, we observe that the light gain factor is notably higher for moderate brightness in WCE images compared to bright areas. As brightness increases, the light gain factor decreases. Leveraging this pattern, the derived light gain factor can effectively enhance high-frequency information in WCE images with moderate brightness. As a result, post-enhanced WCE images exhibit richer details, and this approach effectively achieves adaptive control of the light gain factor.

\begin{figure}[!hptb]
    \begin{center}
    \includegraphics[width=2.1in]{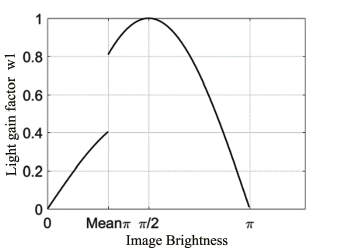}
    \end{center}
\caption{The light gain function reflects the variation law of light gain factor with light intensity.} \label{Fig4}
\end{figure}

\subsubsection{Gradient Gain Factor Construction}
Since traditional USM methods enhance image details using a fixed gain coefficient across different edges, although enhancing the image details, 
the lack of appropriate gain coefficients at the edges leads to edge overshooting. If a greater gain factor is applied to accentuate the details of WCE small intestine villi images, 
while simultaneously utilizing a smaller gain factor at the image edges, edge overshooting is effectively prevented. It is evident that this adaptive implementation of WCE small intestine image enhancement is very necessary. 
To this end, this paper calculates gradient gain factors for specific regions using the gradient gain function. These factors are subsequently integrated into the gain coefficient of refer to (1) to achieve adaptive enhancement of WCE small bowel images. 
The gradient gain function w2 is shown as follows.

\begin{equation} 
	\label{eq5}
w2(x,y)=\begin{cases} e^{\tfrac{In1.5}{0.1}\cdot\triangledown I}-0.5,\triangledown I< 0.1\\ e^{^{3(-\triangledown I+0.1)}},otherwise \end{cases} 
\end{equation}	
Where $(x,y)$ is the image pixel coordinates. $\triangledown I $ is the absolute value of the result after convolution and normalization of the original image with the Laplace operator template in Fig. 5, which represents the edge information of the image.
\begin{figure}[!hptb]
    \begin{center}
    \includegraphics[width=1.1in]{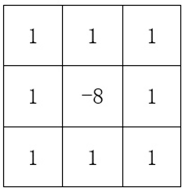}
    \end{center}

\caption{The Laplacian operator template.} \label{Fig5}
\end{figure}

Fig. 6 depicts the gradient gain function. We observe rapid growth in the gradient gain function when the edge information value of the WCE small intestine villi image is less than 0.1. 
This allows the WCE small intestine villi image to quickly obtain a larger gradient gain factor in regions where the edge information value is relatively small. Later, 
the gradient gain factors are introduced into the gain coefficient to achieve the goal of enriching image detail information. 
On the other hand, when the edge information value of the WCE small intestine villi image exceeds 0.1, the slope of the function curve gradually decreases, resulting in a slow reduction of the gradient gain factor. Even in this scenario,
the WCE small intestine villi image still requires a relatively large gradient gain factor. When the edge information value approaches 1, representing the edges in the image, the gradient gain factor is smaller to prevent edge overshoot.

\begin{figure}[!hptb]
    \begin{center}
    \includegraphics[width=2.1in]{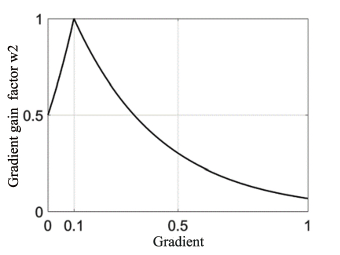}
    \end{center}
\caption{ The gradient gain function reflects the variation pattern of the gradient gain factor with the image gradient.} \label{Fig6}
\end{figure}
\subsection{Enhanced Computing}
To highlight the details of WCE small intestine villi images while preventing excessive noise and overshooting, this paper combines the light gain factor and refer to (1) obtained by refer to (4) and refer to (5), 
so as to realize the adaptive enhancement of WCE small bowel image. The adaptive gain coefficient is shown as follows.

\begin{equation}
	\label{eq6}
	k=\alpha \cdot w1\cdot w2
\end{equation}
Where $\alpha$ is the control maximum parameter, $w1$ is the light gain factor, and $w2$ is the gradient gain factor.

After obtaining the adaptive gain coefficient, we multiply it with the high-frequency component obtained in Section B to obtain the gain high-frequency component of the WCE small intestine image. 
Then, integrate the enhanced component $I$ into the $I$ component of the original input image of the WCE small intestine to strengthen the image's edge. 
Finally, convert from the HSI color space to the RGB color space to obtain the clarified WCE small bowel villi image.

\section{EXPERIMENTS AND RESULTS EVALUATION}
\subsection{Experimental Data Selection}
This paper selected 50 WCE images from the human small intestine as the dataset for image enhancement. 
These images are actual small intestine images of patients collected by capsule endoscopy, aiming to validate the effectiveness of the proposed method. The experiments in this paper were performed on a PC with 8-core Intel i5-8520U, 1.80GHz CPU and 8GB RAM, implemented done using MATLAB 2019a.
\subsection{Evaluation Metrics}
In this paper, Peak Signal-to-noise Ratio (PSNR), Intensity Restricted Average Local Entropy (IRMLE) and Natural Image Quality Evaluator (NIQE) are used for the evaluation of our method.

The PSNR evaluates the noise generated during the image enhancement process. A higher PSNR value of the enhanced image indicates a stronger ability to suppress noise, resulting in better image quality. The PSNR expression is shown as follows.
\begin{equation}
	\label{eq7}
	PSNR=10\times log_{10}\frac{L^{2}}{MSE}
\end{equation}
Where $L$ denotes the maximum intensity in the image. The $MSE$ is the mean square error between the reference image and the image to be measured, and its expression is shown as follows.
\begin{equation}
	\label{eq8}
	MSE=\frac{1}{mn}\sum_{i=0}^{m-1}\sum_{j=0}^{n-1}(I(i,j)-K(i,j))^{^{2}}
\end{equation}
Where $ m\times n$ is the image size. $I$ and $K$ represent the reference image and the pending image respectively.

Compared with natural images, WCE images have limited contrast, and the distribution of their intensities in the whole dynamic range is not uniform. Therefore, the IRMLE proposed in \cite{ref11} is used to evaluate the richness of detail information of WCE images, and the larger the value, the better the detail enhancement effect. The IRMLE expression is shown as follows.
\begin{equation}
	\label{eq9}
	IRMLE=\frac{1}{mn}\sum_{i=1}^{m}\sum_{j=1}^{n}LE(i,j)\cdot \sum \begin{matrix}r_{N}=1\\r_{N}=1/3 \end{matrix}  P(r_{N})
\end{equation}
Where $P(r_{N})$ denotes the probability that the image level is $r_{N}$. $LE(i,j)$ is the entropy of a $9\times9$ window with $(i,j)$ as the pixel center, and the entropy LE is defined as follows.

\begin{equation}
	\label{eq10}
	LE=-\sum_{l=r_{0}}^{r_{L-1}}P(l)\cdot log_{2}P(l)
\end{equation}
Where $P(l)$ represents the probability of the image level being $l$.

NIQE\cite{ref26}is the no-reference evaluation image assessment metric, and the smaller its value, the better the quality. Its quality is expressed as the distance between the NSS feature model and the MVG fitted to the features extracted from the test image, and its expression is shown in refer to (11).
\begin{multline}
	\label{eq11}
	D(v_{1},v_{2},\sum 1 \sum 2 )= \\\sqrt{((v_{1}-v_{2})^{T}(\frac{\sum 1 + \sum 2 }{2})^{-1}(v_{1}-v_{2}))}
\end{multline}
Where $v1,v2,\sum 1,\sum 2 $ represent the mean vector and covariance of the MVG model for natural and test images, respectively.
\subsection{Parameter Variation Experiment}
The parameter $\alpha$ is a constant that controls the enhancement intensity at each point while avoiding excessive enhancement. The specific value of $\alpha$ needs to be determined through experimentation. Fig. 7 illustrates the variation of relevant evaluation parameters when the parameter $\alpha$ takes different values in refer to (6). As the value of $\alpha$ increases, we observe an increase in both the level of detail enhancement and noise increase. This phenomenon indicates that the parameter $\alpha$ is positively correlated with the richness of detail information in WCE small intestine villi images and inversely correlated with the degree of noise. As $\alpha$ increases, the NIQE value initially decreases and then increases. When $\alpha$ is set to 4, the NIQE value reaches its minimum at 3.5015, while the PSNR value is 33.1341, and the IRMLE value is 2.5345. Overall, the image quality is at its best. Effectively maintaining the original small intestine structure, enhancing the clarity of small intestine villi details, and preventing excessive noise and edge artifacts. In this paper, the value of $\alpha$ is selected as 4.

\begin{figure}[H]
	
	\centering
	\begin{subfigure}{0.45\linewidth}
		\centering
		\includegraphics[width=0.95\linewidth]{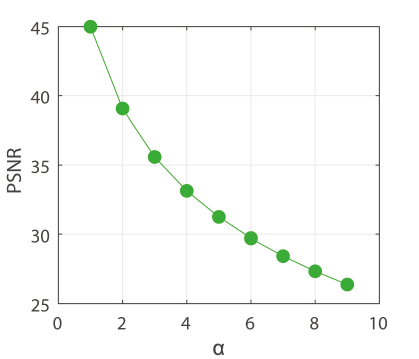}
		\caption{PSNR}
		\label{Fig71}%
	\end{subfigure}
	\centering
	\begin{subfigure}{0.46\linewidth}
		\centering
		\includegraphics[width=0.95\linewidth]{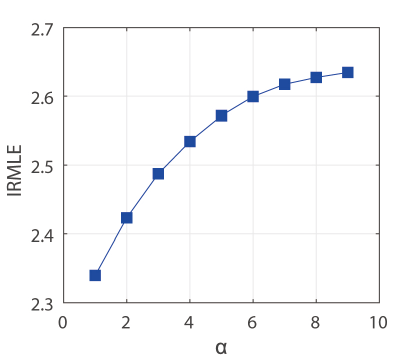}
		\caption{IRMLE}
		\label{Fig72}
	\end{subfigure}
\end{figure}
\begin{figure}[H]\ContinuedFloat
	\centering
	\begin{subfigure}{0.46\linewidth}
		\centering
		\includegraphics[width=0.95\linewidth]{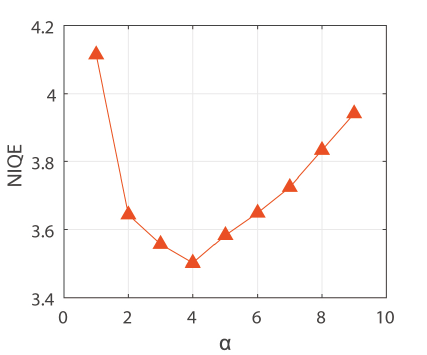}
		\caption{NIQE}
		\label{Fig73}%
	\end{subfigure}

	\caption{ The parameter $\alpha$ takes different values, ranging from 1 to 9, to evaluate the enhancement results of WCE small intestine images.  (a) shows the trend of PSNR changes, (b) illustrates the trend of IRMLE changes, and (c) portrays the trend in NIQE values.}
	\label{Fig7}
\end{figure}

When the value of $\alpha$ is determined to be 4, the influence of the lightness gain factor w1 and the gradient gain factor w2 on the proposed method is further discussed. $w1(x, y)$ and $w2(x, y)$ are adaptive functions at each pixel point $(x, y)$ on the image, and their values change depending on the pixel point $(x, y)$. Therefore, experiments were conducted on three representative WCE small intestine villi images, as shown in Fig. 8(a), (b), and(c) to examine the effects of these factors on enhancing small intestine villi images. 

\begin{figure}[!htbp]
	\centering
	\begin{subfigure}{0.28\linewidth}
		\centering
		\includegraphics[width=0.95\linewidth]{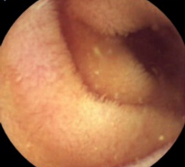}
		\caption{original}
		\label{Fig81}
	\end{subfigure}
	\centering
	\begin{subfigure}{0.28\linewidth}
		\centering
		\includegraphics[width=0.95\linewidth]{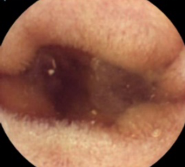}
		\caption{original}
		\label{Fig82}
	\end{subfigure}
	\centering
	\begin{subfigure}{0.28\linewidth}
		\centering
		\includegraphics[width=0.95\linewidth]{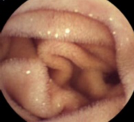}
		\caption{original}
		\label{Fig83}
	\end{subfigure}

	\caption{ Parameter $w1$ and $w2$ ablation experiments. (a), (b), and (c) represent the original WCE small intestine images.}
	\label{Fig8}
\end{figure}

\begin{table}[!hptb]
	\newcommand{\tabincell}[2]{\begin{tabular}{@{}#1@{}}#2\end{tabular}}
	\centering
	\caption{Evaluation index values of WCE small intestine enhancement results in ablation experiments}
	\setlength{\tabcolsep}{3pt}
	\renewcommand{\arraystretch}{1.2}
	\begin{tabular}{ccccc}  
	  \toprule
	  \multicolumn{2}{c}{The ablation part} &
	 PSNR  & IRMLE & NIQE \\ \hline
	 \multirow{3}{*}{\tabincell{c}{Light gain factor \\ (w1)}}& image (a)  & 29.7378 & \textbf{2.1777} & 3.7072\\
	 & image (b)  & 29.1702 & \textbf{1.8249} & 4.9491 \\ 
	 & image (c)  & 27.7551 & \textbf{2.7342} & 3.7062 \\ \cline{2-5}
	 \multirow{3}{*}{\tabincell{c}{Gradient gain factor \\ (w2)}}& image (a) & 32.1409&2.1517&	3.6763\\
	 & image (b)&31.7118&1.8097&4.9391 \\ 
	 & image (c)&30.2093&2.7293&3.6369\\ \cline{2-5}
	 \multirow{3}{*}{w1+w2}& image (a) & \textbf{34.2702}&2.0775&\textbf{3.6381}\\
	 & image (b)&\textbf{33.7453}&1.7136&\textbf{4.8975}\\ 
	 & image (c)&\textbf{32.2204}&2.6868&\textbf{3.4683}\\ 
	  \bottomrule
	\end{tabular}
	
	\label{tab1}
  \end{table}   
Table 1 provides the evaluation metric values for the enhanced results in Fig. 8. When considering only the lightness gain factor, we observe that it has the best IRMLE value compared to the other two cases, but it exhibits the worst PSNR and NIQE values. This suggests that when only the lightness gain factor is considered, although the noise in dark areas is suppressed, there is an excessive enhancement of edge details in WCE small intestine villi images. On the other hand, when considering only the gradient gain factor, although the IRMLE value decreases compared to when considering only the lightness gain factor, the PSNR and NIQE values show some improvement. However, compared to the case of jointly considering both the lightness gain factor and the gradient gain factor, the PSNR and NIQE values are still relatively poorer. This suggests that the edge details of WCE small intestine villi images are adaptively enhanced, but there is an excessively enhanced noise in dark areas. When jointly considering both the lightness gain factor and the gradient gain factor, the PSNR and NIQE values are optimal. This phenomenon indicates effective enhancement of edge details in WCE small intestine villi images while suppressing excessive noise, resulting in the best visual effect.

As shown in Fig. 9, the highlighted results of small intestine villi clarification in WCE are presented. The parameters used for enhancement are derived from the experimental results mentioned earlier. In Fig. 9(e)-(h), the villi within the blue boxes are notably clearer compared to the original images. The method employed in this paper incorporates adaptive gradient enhancement, resulting in enhanced villi without the occurrence of edge artifacts. In addition, the green boxes in Fig. 9(e)-(h) represent enhanced dark areas. The original images in Fig. 9(a)-(d) exhibit limited information and are filled with noise in these dark regions. However, the method introduced in this paper employs light gain factors to suppress noise in dark areas, as evident in the enhanced results. Furthermore, it is apparent from Fig. 9 that the proposed method effectively highlights the small intestine villi, which provides medical professionals with improved image clarity for diagnosing medical conditions.
\begin{figure}[!hptb]
	\centering
	\begin{subfigure}{0.24\linewidth}
		\centering
		\includegraphics[width=0.95\linewidth]{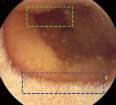}
		\caption{original}
		\label{Fig91}%
	\end{subfigure}
	\centering
	\begin{subfigure}{0.24\linewidth}
		\centering
		\includegraphics[width=0.95\linewidth]{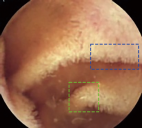}
		\caption{original}
		\label{Fig92}%
	\end{subfigure}
	\centering
	\begin{subfigure}{0.24\linewidth}
		\centering
		\includegraphics[width=0.95\linewidth]{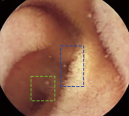}
		\caption{original}
		\label{Fig93}%
	\end{subfigure}
	\centering
	\begin{subfigure}{0.24\linewidth}
		\centering
		\includegraphics[width=0.95\linewidth]{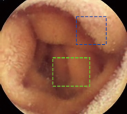}
		\caption{original}
		\label{Fig94}%
	\end{subfigure}
	\centering
	\begin{subfigure}{0.24\linewidth}
		\centering
		\includegraphics[width=0.95\linewidth]{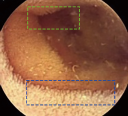}
		\caption{proposed}
		\label{Fig95}%
	\end{subfigure}
	\centering
	\begin{subfigure}{0.24\linewidth}
		\centering
		\includegraphics[width=0.95\linewidth]{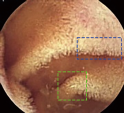}
		\caption{proposed}
		\label{Fig96}%
	\end{subfigure}
	\centering
	\begin{subfigure}{0.24\linewidth}
		\centering
		\includegraphics[width=0.95\linewidth]{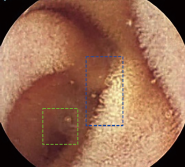}
		\caption{proposed}
		\label{Fig97}%
	\end{subfigure}
	\centering
	\begin{subfigure}{0.24\linewidth}
		\centering
		\includegraphics[width=0.95\linewidth]{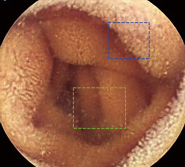}
		\caption{proposed}
		\label{Fig98}%
	\end{subfigure}
	\caption{ WCE small intestinal villi highlighting results. (a)-(d) are the original input images of WCE small intestine, and (e)-(h) are the corresponding results after enhancement using the method of this paper, respectively. The blue box is selected for the villi-rich area of the enhanced small intestine, and the green box is selected for the dark area of the enhanced small intestine.  }
	\label{Fig9}
\end{figure}

\begin{figure}[!hptb]
	\centering
	\begin{subfigure}{0.46\linewidth}
		\centering
		\includegraphics[width=0.95\linewidth]{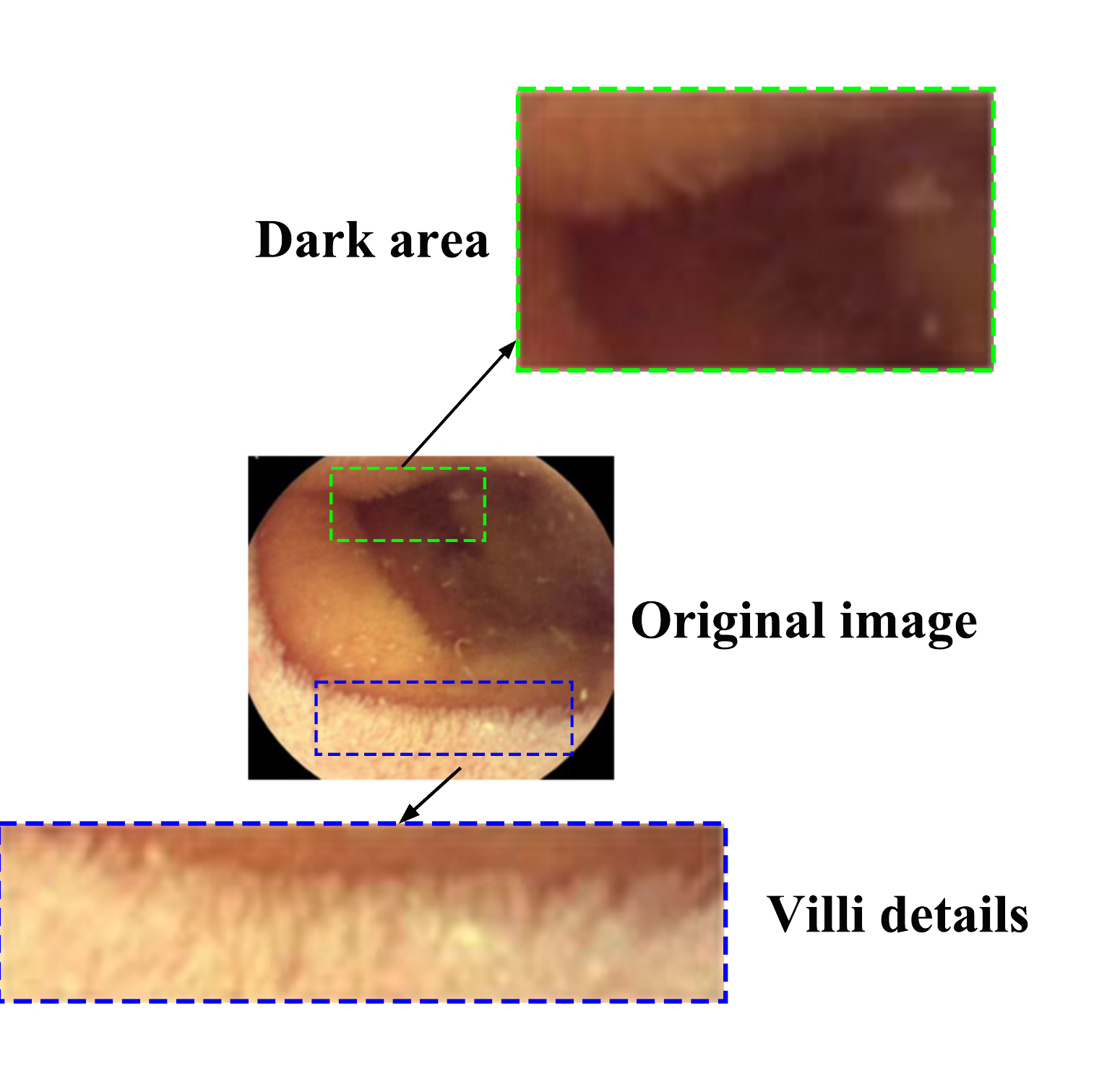}
		\caption{original}
		\label{Fig101}%
	\end{subfigure}
	\centering
	\begin{subfigure}{0.46\linewidth}
		\centering
		\includegraphics[width=0.95\linewidth]{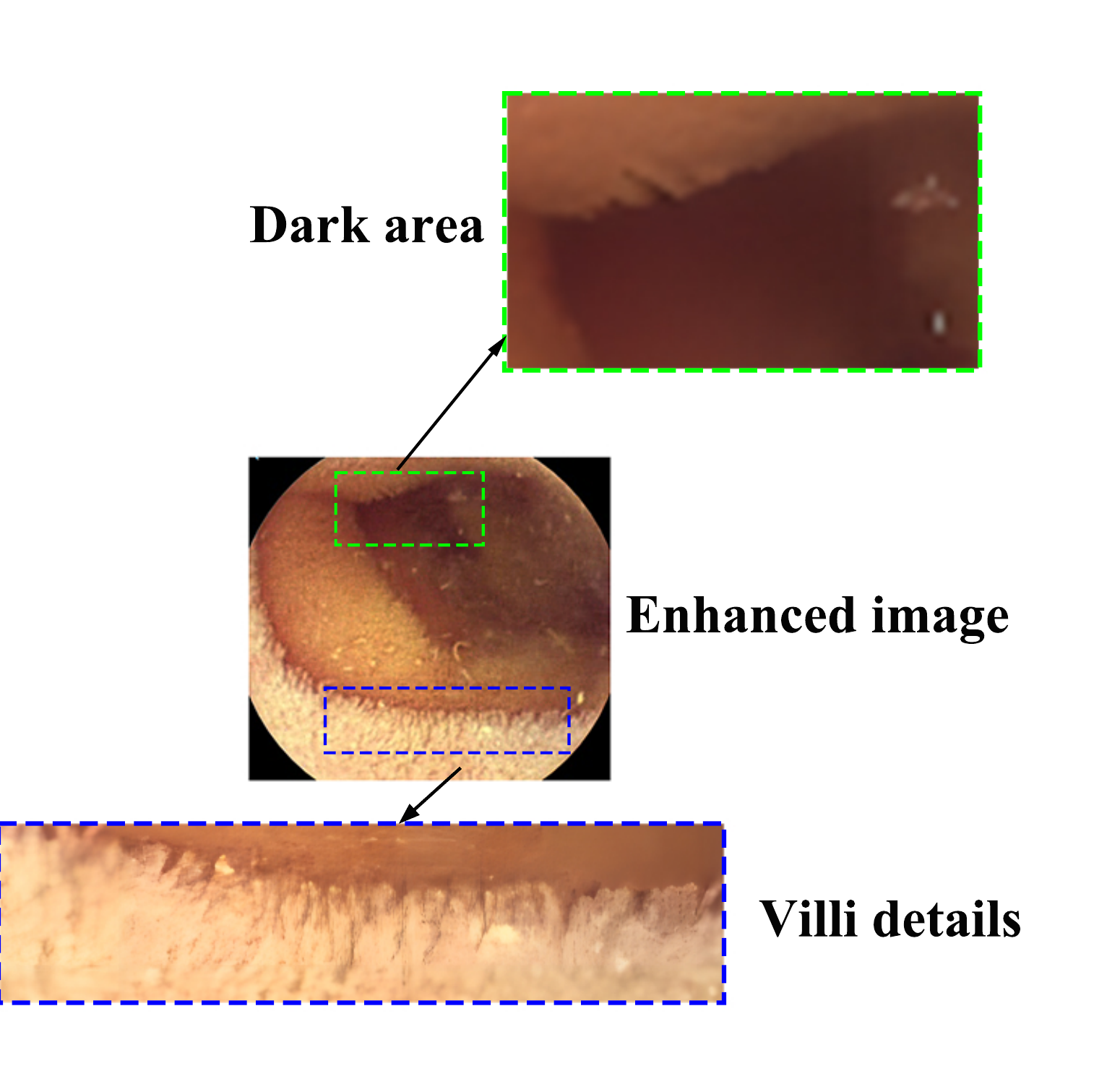}
		\caption{proposed}
		\label{Fig102}%
	\end{subfigure}
	\centering
	\begin{subfigure}{0.46\linewidth}
		\centering
		\includegraphics[width=0.95\linewidth]{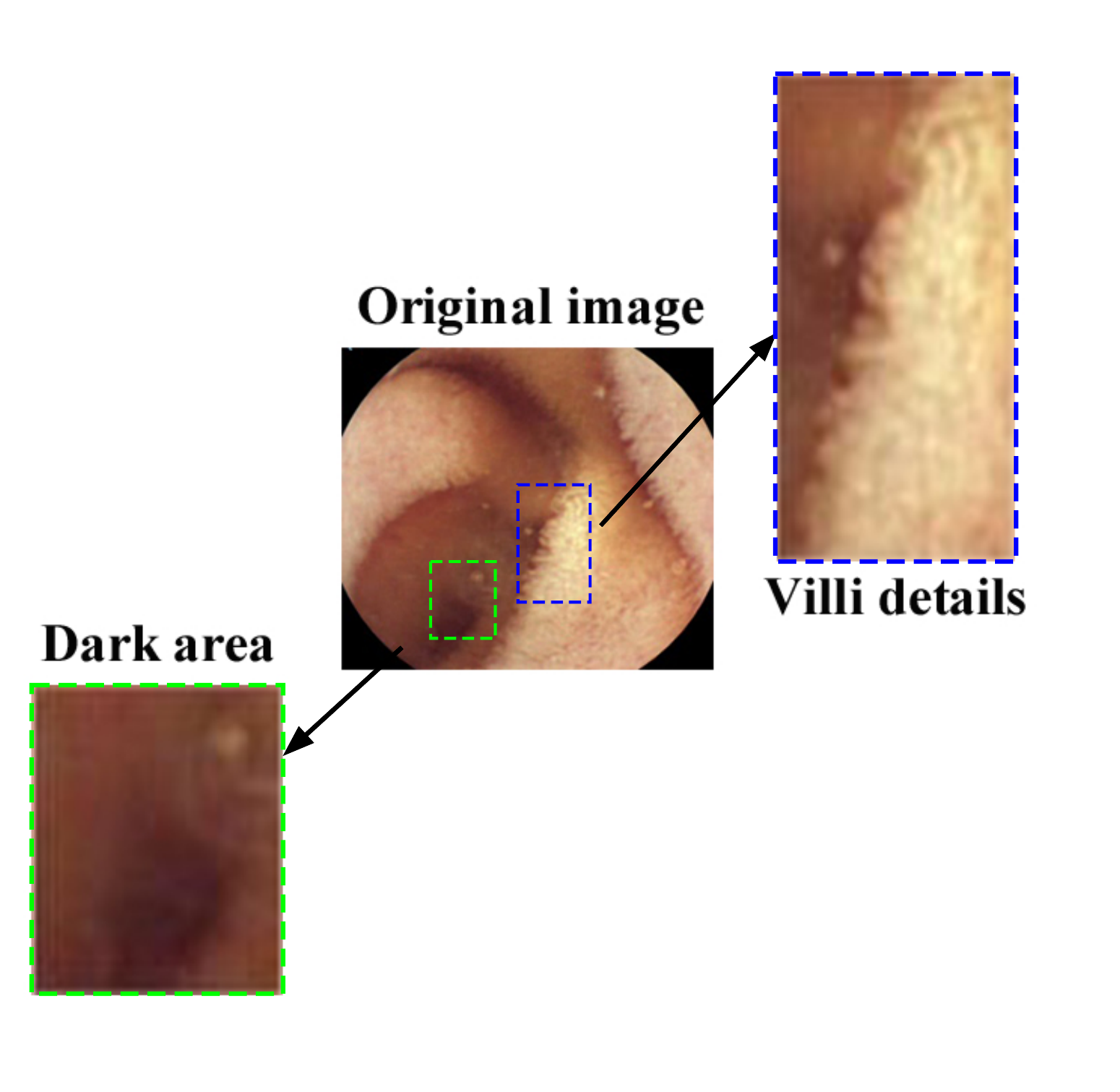}
		\caption{original}
		\label{Fig103}%
	\end{subfigure}
	\centering
	\begin{subfigure}{0.46\linewidth}
		\centering
		\includegraphics[width=0.95\linewidth]{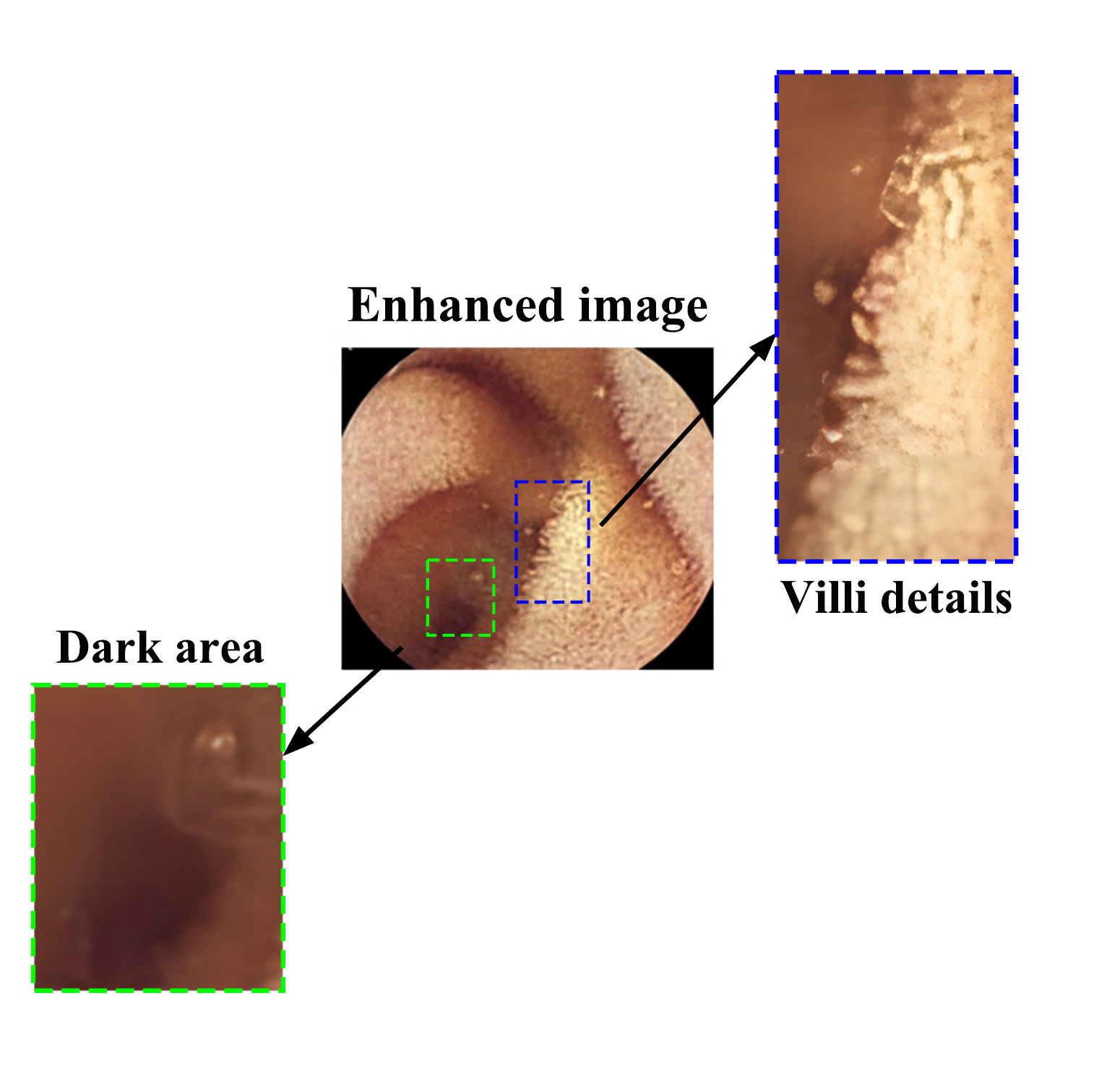}
		\caption{proposed}
		\label{Fig104}%
	\end{subfigure}
	\caption{ Selecting (a), (c) from the original small intestine images in Fig. 9, and (e), (g) from the enhanced small intestine images in Fig. 9, details were magnified. Fig. 10(a), (c) respectively corresponds to Fig. 9(a), (c) and Fig. 10(b), (d) respectively corresponds to Fig. 9(e), (g).}
	\label{Fig10}
\end{figure}
\vspace{-10pt}

To facilitate closer examination, Fig. 10 depicts selected examples using the small intestine's original input images from Fig. 9(a), (c) and the enhanced results with highlighted villi from Fig. 9(e), (g), allowing for a more detailed observation  of small intestine villi details.

\subsection{Experimental Comparison Analysis}
In this paper, 50 WCE small intestine images were enhanced using HCUM \cite{ref11}, NLUM\cite{ref12}, UMGF\cite{ref13} and the method of this paper. Table 2 shows the mean values of PSNR, IRMLE and NIQE after processing by the above methods, where the best objective values are shown in bold. Fig. 11(a), (b), and(c) shows the PSNR, IRMLE and NIQE values for the WCE small intestine images after each method individually processes them. In Table 2, although HCUM shows a decent IRMLE value, its PSNR and NIQE values are poor, indicating that using this method to enhance the details of WCE small intestine images may result in excessive noise enhancement. Additionally, the structure of the enhanced WCE small intestine images is altered. In Table 2, the IRMLE value of UMGF decreased compared to the original image. The WCE small intestine villus images enhanced using this method exhibit excessive brightness enhancement, resulting in detail enhancement occurring in excessively bright and excessively dark areas. Fig. 12(b), (g), and (l) shows the results after HCUM enhancement, Fig. 12(d), (i), and (n) shows the results after UMGF enhancement. We can observe that the WCE small intestine villi images enhanced by HCUM and UMGF exhibit excessive noise enhancement, structural changes, and edge overshooting. In Table 2, NLUM exhibits good PSNR, IRMLE, and NIQE values. Additionally, the NLUM enhancement results in Fig. 12(c), (h), and (m) reveal that the structure of the WCE small intestine villi image is well-preserved without the occurrence of edge overshooting. However, there is an issue of excessive noise enhancement in the dark areas of the WCE small intestine image after enhancement.

Compared to other methods, our approach yields the highest average PSNR values for the enhanced WCE small intestine villus images, along with significant improvements in IRMLE and NIQE values. This suggests that our method effectively enhances the details of WCE small intestine villus images while suppressing the increase in noise, resulting in a visually pleasing enhancement. Moreover, as shown in Fig. 12(e), (j), and (o) depicting the results of this method, it also prevents the occurrence of edge artifacts.

\begin{table}[!hptb]
	\centering
	\caption{Average value of evaluation indicators using different methods}
	\setlength{\tabcolsep}{15pt}
	\begin{tabular} {l *{3}{S[table-format=2.4]}}  
	  \toprule
	  Method & {PSNR} & {IRMLE} & {NIQE} \\
	  \midrule
	  original & {--} & 1.6640 & 4.7734 \\
	  HCUM & 19.8565 & \textbf{ 2.4620} & 5.2023 \\
	  NLUM & 31.3513 & 2.0271 & \textbf{ 3.0818} \\
	  UMGF & 19.7736 & 1.6555 & 3.9417 \\
	  proposed & \textbf{34.4204} & 1.8743 & 3.2532 \\
	  \bottomrule
	\end{tabular}
	
	\label{tab2}
  \end{table}    

\begin{figure}[h]
	\centering
	\begin{subfigure}{0.44\linewidth}
		\centering
		\includegraphics[width=0.95\linewidth]{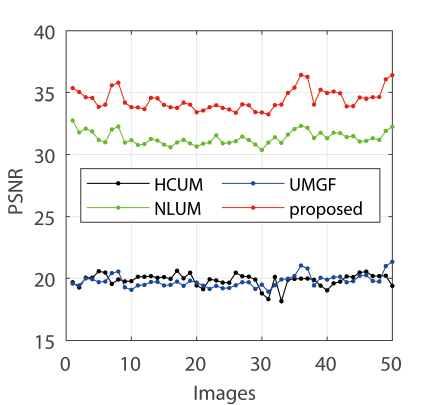}
		\caption{PSNR}
		\label{Fig111}
	\end{subfigure}
	\centering
	\begin{subfigure}{0.47\linewidth}
		\centering
		\includegraphics[width=0.95\linewidth]{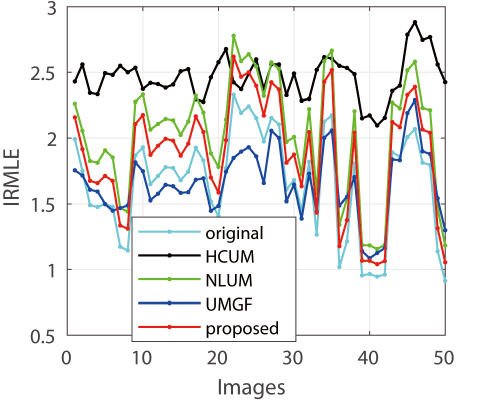}
		\caption{IRMLE}
		\label{Fig112}
	\end{subfigure}
	\centering
	\begin{subfigure}{0.46\linewidth}
		\centering
		\includegraphics[width=0.95\linewidth]{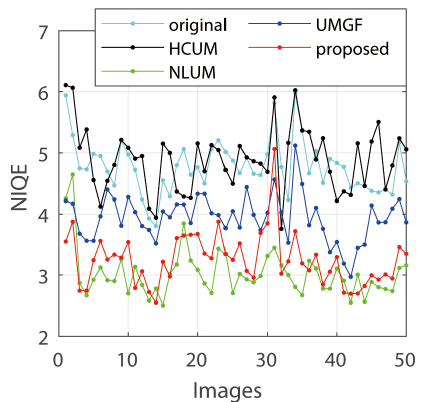}
		\caption{NIQE}
		\label{Fig113}
	\end{subfigure}

	\caption{ Enhance 50 sets of small intestine images using different methods and evaluate the enhancement results using assessment metrics.}
	\label{Fig11}
\end{figure}

\begin{figure}[!hptb]
	\centering
	\begin{subfigure}{0.19\linewidth}
		\centering
		\includegraphics[width=0.95\linewidth]{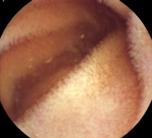}
		\caption{original}
		\label{Fig121}
	\end{subfigure}
	\centering
	\begin{subfigure}{0.19\linewidth}
		\centering
		\includegraphics[width=0.95\linewidth]{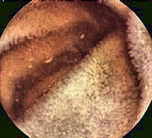}
		\caption{HCUM}
		\label{Fig122}
	\end{subfigure}
	\centering
	\begin{subfigure}{0.19\linewidth}
		\centering
		\includegraphics[width=0.95\linewidth]{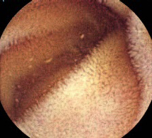}
		\caption{NLUM}
		\label{Fig123}
	\end{subfigure}
	\centering
	\begin{subfigure}{0.19\linewidth}
		\centering
		\includegraphics[width=0.95\linewidth]{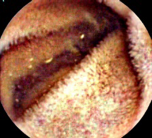}
		\caption{UMGF}
		\label{Fig124}
	\end{subfigure}
	\centering
	\begin{subfigure}{0.19\linewidth}
		\centering
		\includegraphics[width=0.95\linewidth]{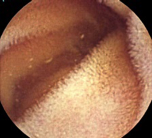}
		\caption{proposed}
		\label{Fig125}
	\end{subfigure}

	\centering
	\begin{subfigure}{0.19\linewidth}
		\centering
		\includegraphics[width=0.95\linewidth]{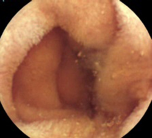}
		\caption{original}
		\label{Fig126}
	\end{subfigure}
	\centering
	\begin{subfigure}{0.19\linewidth}
		\centering
		\includegraphics[width=0.95\linewidth]{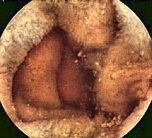}
		\caption{HCUM}
		\label{Fig127}%
	\end{subfigure}
	\centering
	\begin{subfigure}{0.19\linewidth}
		\centering
		\includegraphics[width=0.95\linewidth]{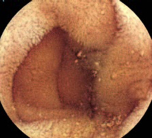}
		\caption{NLUM}
		\label{Fig128}
	\end{subfigure}
	\centering
	\begin{subfigure}{0.19\linewidth}
		\centering
		\includegraphics[width=0.95\linewidth]{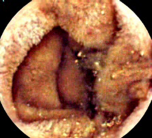}
		\caption{UMGF}
		\label{Fig129}
	\end{subfigure}
	\centering
	\begin{subfigure}{0.19\linewidth}
		\centering
		\includegraphics[width=0.95\linewidth]{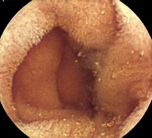}
		\caption{proposed}
		\label{Fig1210}
	\end{subfigure}
	
	\centering
	\begin{subfigure}{0.19\linewidth}
		\centering
		\includegraphics[width=0.95\linewidth]{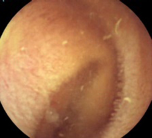}
		\caption{original}
		\label{Fig1211}
	\end{subfigure}
	\centering
	\begin{subfigure}{0.19\linewidth}
		\centering
		\includegraphics[width=0.95\linewidth]{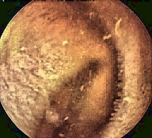}
		\caption{HCUM}
		\label{Fig1212}
	\end{subfigure}
	\centering
	\begin{subfigure}{0.19\linewidth}
		\centering
		\includegraphics[width=0.95\linewidth]{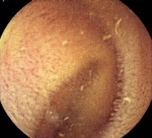}
		\caption{NLUM}
		\label{Fig1213}
	\end{subfigure}
	\centering
	\begin{subfigure}{0.19\linewidth}
		\centering
		\includegraphics[width=0.95\linewidth]{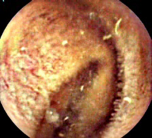}
		\caption{UMGF}
		\label{Fig1214}
	\end{subfigure}
	\centering
	\begin{subfigure}{0.19\linewidth}
		\centering
		\includegraphics[width=0.95\linewidth]{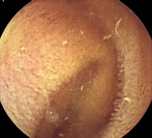}
		\caption{proposed}
		\label{Fig1215}
	\end{subfigure}

	\caption{ Three different sets of representative images of small intestine were selected and enhanced using HCUM, NLUM, UMGF, and the methods in this paper, respectively.}
	\label{Fig12}
\end{figure}

To further verify the effectiveness of the method in this paper, the enhancement results of the above HCUM, NLUM and UMGF methods were denoised using gaussian filters. Table 3 shows the mean values of PSNR, IRMLE, and NIQE after the denoising process, where the best objective values are shown in bold. Fig. 13(a), (b), and (c) shows the denoised PSNR, IRMLE and NIQE values of each WCE small bowel image. Fig. 14 shows the results of HCUM, NLUM and UMGF enhancement followed by denoising process. In Fig. 14, after denoising, the enhancement results of HCUM, NLUM, and UMGF exhibit a noticeable reduction in noise. Compared to Table 2, the PSNR values in Table 3 have improved, and the IRMLE values have decreased. The mentioned alterations suggest that WCE small intestine villi images enhanced through HCUM, NLUM, and UMGF contain excessive noise in the details, which affects both the enhancement of details and the preservation of the structure. Moreover, in Table 3 compared to Table 2, the NIQE value of HCUM decreases after denoising, indicating an excessive amount of noise in the enhanced dark areas. The quality of the enhancement results improves after denoising. The NIQE values of NLUM and UMGF increase after denoising, indicating an excessive enhancement of details. Although there is a slight smoothing of details during noise removal, as seen in Fig. 14, the problem of excessive enhancement still persists. In Table 3, the PSNR and NIQE values after denoising with HCUM, NLUM, and UMGF still fall short of the method presented in this paper. This indicates that our method is capable of effectively emphasizing the details of small intestine villi while suppressing noise and preventing over-enhancement. Furthermore, the clarity-enhanced small intestine villus images generated using our method exhibit good visual quality. It is crucial to highlight that our method's enhancement performance is not satisfactory for particularly dark areas, as evidenced by the experimental results in Fig. 10. The main reason for this is the severe lack of information in those regions of the images.

\begin{table}[!hptb]
	\centering
	\caption{Average value of evaluation indicators using different methods after denoising}
	\setlength{\tabcolsep}{15pt}
	\begin{tabular}{l *{3}{S[table-format=2.4]}}  
	  \toprule
	  Method & {PSNR} & {IRMLE} & {NIQE} \\
	  \midrule
	  original & {--} & 1.6640 & 4.7734 \\
	  HCUM & 21.0051 & \textbf{2.3990} & 3.9505 \\
	  NLUM & 31.9147 & 1.9353 & 4.5599 \\
	  UMGF & 20.0433 & 1.6610 & 5.1426 \\
	  proposed & \textbf{34.4204} & 1.8743 & \textbf{ 3.2532} \\
	  \bottomrule
	\end{tabular}
	
	\label{tab3}
  \end{table}

  \begin{figure}[H]
	
	\centering
	\begin{subfigure}{0.45\linewidth}
		\centering
		\includegraphics[width=0.95\linewidth]{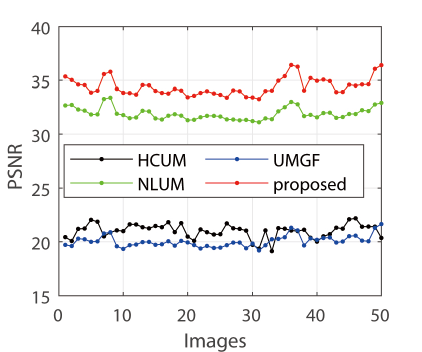}
		\caption{PSNR}
		\label{Fig131}%

	\end{subfigure}
	\centering
	\begin{subfigure}{0.46\linewidth}
		\centering
		\includegraphics[width=0.95\linewidth]{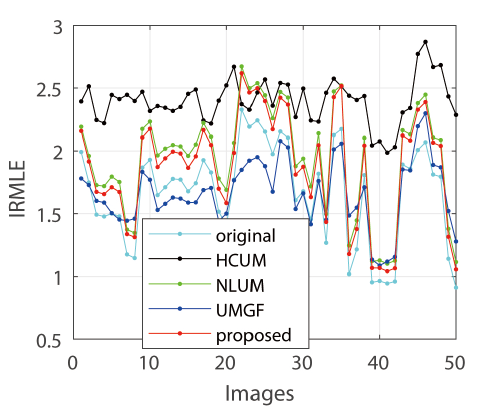}
		\caption{IRMLE}
		\label{Fig132}
	\end{subfigure}
\end{figure}
\begin{figure}[H]\ContinuedFloat
	\centering
	\begin{subfigure}{0.46\linewidth}
		\centering
		\includegraphics[width=0.95\linewidth]{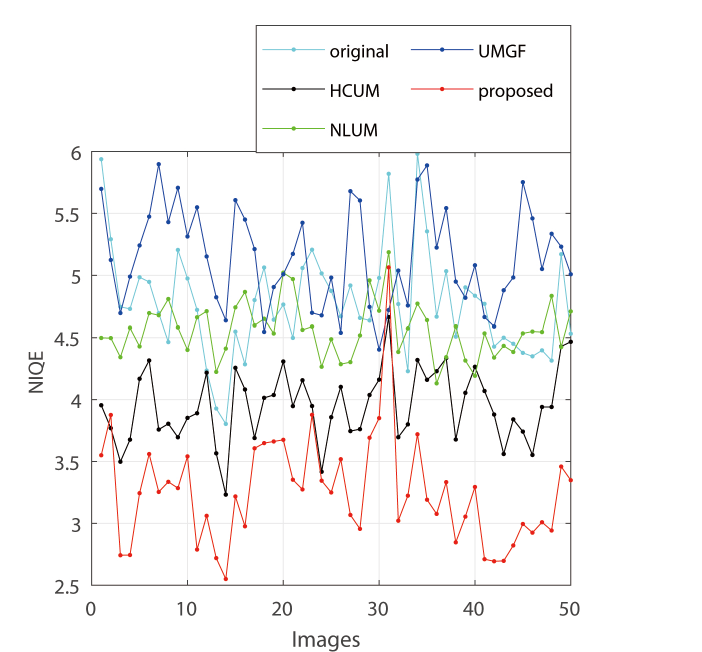}
		\caption{NIQE}
		\label{Fig133}%
	\end{subfigure}

	\caption{ Enhancement of 50 sets of denoised small intestine images using different methods and evaluation of the enhancement results.}
	\label{Fig13}
\end{figure}

\begin{figure}[!hptb]
	\centering
	\begin{subfigure}{0.19\linewidth}
		\centering
		\includegraphics[width=0.95\linewidth]{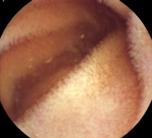}
		\caption{original}
		\label{Fig141}
	\end{subfigure}
	\centering
	\begin{subfigure}{0.19\linewidth}
		\centering
		\includegraphics[width=0.95\linewidth]{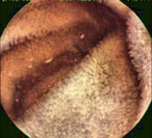}
		\caption{HCUM}
		\label{Fig142}
	\end{subfigure}
	\centering
	\begin{subfigure}{0.19\linewidth}
		\centering
		\includegraphics[width=0.95\linewidth]{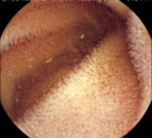}
		\caption{NLUM}
		\label{Fig143}
	\end{subfigure}
	\centering
	\begin{subfigure}{0.19\linewidth}
		\centering
		\includegraphics[width=0.95\linewidth]{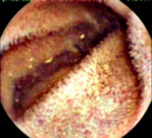}
		\caption{UMGF}
		\label{Fig144}
	\end{subfigure}
	\centering
	\begin{subfigure}{0.19\linewidth}
		\centering
		\includegraphics[width=0.95\linewidth]{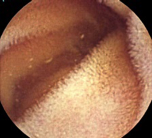}
		\caption{proposed}
		\label{Fig145}
	\end{subfigure}

	\centering
	\begin{subfigure}{0.19\linewidth}
		\centering
		\includegraphics[width=0.95\linewidth]{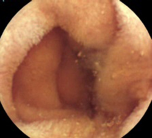}
		\caption{original}
		\label{Fig146}
	\end{subfigure}
	\centering
	\begin{subfigure}{0.19\linewidth}
		\centering
		\includegraphics[width=0.95\linewidth]{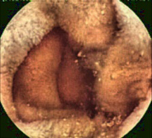}
		\caption{HCUM}
		\label{Fig147}
	\end{subfigure}
	\centering
	\begin{subfigure}{0.19\linewidth}
		\centering
		\includegraphics[width=0.95\linewidth]{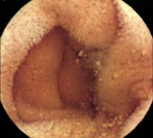}
		\caption{NLUM}
		\label{Fig148}%
	\end{subfigure}
	\centering
	\begin{subfigure}{0.19\linewidth}
		\centering
		\includegraphics[width=0.95\linewidth]{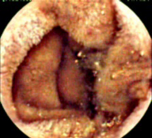}
		\caption{UMGF}
		\label{Fig149}%
	\end{subfigure}
	\centering
	\begin{subfigure}{0.19\linewidth}
		\centering
		\includegraphics[width=0.95\linewidth]{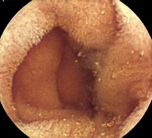}
		\caption{proposed}
		\label{Fig1410}%
	\end{subfigure}
	
	\centering
	\begin{subfigure}{0.19\linewidth}
		\centering
		\includegraphics[width=0.95\linewidth]{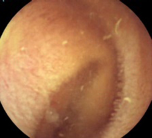}
		\caption{original}
		\label{Fig1411}%
	\end{subfigure}
	\centering
	\begin{subfigure}{0.19\linewidth}
		\centering
		\includegraphics[width=0.95\linewidth]{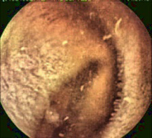}
		\caption{HCUM}
		\label{Fig1412}%
	\end{subfigure}
	\centering
	\begin{subfigure}{0.19\linewidth}
		\centering
		\includegraphics[width=0.95\linewidth]{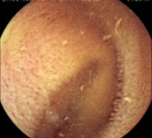}
		\caption{NLUM}
		\label{Fig1413}%
	\end{subfigure}
	\centering
	\begin{subfigure}{0.19\linewidth}
		\centering
		\includegraphics[width=0.95\linewidth]{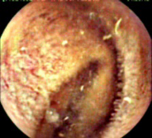}
		\caption{UMGF}
		\label{Fig1414}%
	\end{subfigure}
	\centering
	\begin{subfigure}{0.19\linewidth}
		\centering
		\includegraphics[width=0.95\linewidth]{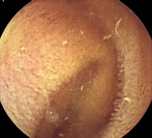}
		\caption{proposed}
		\label{Fig1415}%
	\end{subfigure}

	\caption{ Three sets of representative small intestine image collections were selected and enhanced using HCUM, NLUM, UMGF, and the method presented in this paper, followed by denoising, respectively.}
	\label{Fig14}
\end{figure}

\section{	CONCLUSION}
This paper presents an image enhancement method for improving the clarity of small intestine villi. This approach aims to address issues such as blurriness, detail loss, edge artifacts, and noise amplification in capsule endoscopy images capturing the villi-rich small intestine. Experimental results confirm that the proposed method leads to a 12.63 improvement in the IRMLE compared to the original images, a 45.47 increase in PSNR relative to classical enhancement algorithms, and a 31.84 reduction in NIQE.

Future work will build upon the findings of this paper, focusing on the enhancement of villi clarity in extremely dark regions in WCE images.





\end{document}